\documentclass[11pt]{amsart}
\usepackage{times,amsmath,amsbsy,amssymb}
\usepackage[pdftex,colorlinks]{hyperref}
\usepackage{color}
\usepackage{float}
\usepackage{subfig}
\usepackage[margin=1in]{geometry}
\usepackage{multirow}
\usepackage{bm}
\usepackage{dsfont}
\usepackage{cite}
\usepackage[utf8]{inputenc}
\usepackage{graphicx}
\newtheorem{theorem}{Theorem}[section]

\newtheorem{remark}[theorem]{Remark}
\usepackage{algorithm}
\usepackage{algpseudocode}
\numberwithin{equation}{section} \numberwithin{table}{section}

\DeclareGraphicsExtensions{.png,.jpg,.pdf,.gif,.eps}
\graphicspath{{./sin/}{./single_layer_mnist/}{./NNETS/}}

\usepackage[colorinlistoftodos,prependcaption,textsize=footnotesize]{todonotes}
\newcommand{\SC}[1]
{
\todo[linecolor=cyan,backgroundcolor=cyan!25,bordercolor=cyan] 
{#1}}

\renewcommand{\SC}[1]{}

\title{Neural networks with trainable matrix activation functions}
\author{Zhengqi Liu}
\address{Department of Mathematics, The Pennsylvania State University, University Park, PA 16802, USA}
  \email{zbl5196@psu.edu}

\author{Shuhao Cao}
\address{School of Science and Engineering, University of Missouri-Kansas City, Kansas City, MO 64110, USA}
  \email{scao@umkc.edu}

\author{Yuwen Li}
\address{School of Mathematical Sciences, Zhejiang University, Hangzhou, Zhejiang 310058, China}
  \email{liyuwen@zju.edu.cn}

\author{Ludmil Zikatanov}
\address{Department of Mathematics, The Pennsylvania State University, University Park, PA 16802, USA}
    \email{ltz1@psu.edu}
    
\date{\today}

\clearpage
\begin{document}

\begin{abstract}
The training process of neural networks usually optimize weights and bias parameters of linear transformations, while nonlinear activation functions are pre-specified and fixed. This work develops a systematic approach to constructing matrix-valued activation functions whose entries are generalized from ReLU. The activation is based on matrix-vector multiplications using only scalar multiplications and comparisons. The proposed activation functions depend on parameters that are trained along with the weights and bias vectors. Neural networks based on this approach are simple and efficient and are shown to be robust in numerical experiments.
\end{abstract}
\maketitle

\section{Introduction}

In recent decades, deep neural networks (DNNs) have achieved significant successes in many fields such as computer vision and natural language processing\cite{Voulodimos2018,DBLP:journals/corr/abs-1807-10854}. The DNN surrogate model is constructed using recursive composition of linear transformations and nonlinear activation functions. 
The nonlinear activation functions are essential in provide universal approximation, and 
problem-appropriate choices of them are vital to the model's performance.

In the original universal approximation~\cite{funahashi1989approximate,cybenko1989approximation}, sigmoid is used due to its property converging to $1$ and $0$ as the input goes to $\pm \infty$ and continuously changing within. However, in the modern application where the neural network layers are composed to be deeper and deeper while the whole community shifted from fp64 to fp32, sigmoid activation suffers from the ``vanishing gradient''~\cite{2001vanishing} during the training process, as the chain rule multiplies multiple small gradient resulting from sigmoid and eventually causes numerical underflow~\cite{you2017large}.

In practice, Rectified Linear Unit (ReLU) is one of the most popular activation functions due its simplicity, efficiency. Moreover, it resolves the vanishing gradient problem completely allowing large DNN stacked with up to hundreds of layers such as the ones in \cite{HeZRS15}. Nevertheless, a key drawback of ReLU is the ``dying ReLU'' problem~\cite{Lu2020}, where if during training a neuron's ReLU activation becomes 0, then under certain circumstance it may never get activated again to output a nonzero value.

\SC{Dying relu and vanishing gradient are two different things, I changed the writing here.}

Several simple modifications are proposed to address this problem and achieved certain level of successes, e.g., the simple Leaky ReLU, and
Piecewise Linear Unit (PLU) \cite{plu}, Softplus  \cite{softplus}, Exponential Linear Unit (ELU) \cite{elu}, Scaled Exponential Linear Unit (SELU) \cite{selu}, and Gaussian Error Linear Unit (GELU) \cite{gelu}.
\SC{Some citations are needed, for example, why Leaky ReLU is good: ``Leaky-RELU prevents negative-activated neurons from having a zero-gradient which can prevent neurons from `dying'.''}

 
Although the aforementioned activation functions are shown to be competitive in benchmark tests, they are still fixed nonlinear functions. In a DNN structure, it is often hard to determine a priori the optimal activation function for a specific application. Empirically, there has been a community effort in search for a better activation~\cite{ramachandran2017searching}, and currently the consensus is that GELU works well in large models such as GPT~\cite{radford2019language} as it avoids the vanishing gradient, dying ReLU, as well as the shattered gradient problem~\cite{balduzzi2017shattered}. In general, GELU-like activations provides gradient in the negative regime to stop neurons ``dying'' while bounding how far into the negative regime activations are able to have an effect, and this allows for a better cross-layer training procedure.  \SC{I think this part is worth to be expanded, why it is hard to determine? Is there an example? I wrote something here I think relevant.} 

However, all the activations mentioned above are \emph{pointwise}, where this pointwise-ness refers the fact that the activation are scalar functions that only use a single component-wise value to determine its output given a tensorial input. In this paper, we shall generalize these activation functions and introduce a mechanism to create matrix-valued activation functions, which is trainable to control gradient magnitude in a data-adaptive fashion. The effectiveness of the proposed method is validated using function approximation examples and well-known benchmark datasets such as \texttt{MNIST} and \texttt{CIFAR-10}. There are a few classical works on adaptively tuning of parameters in the training process, e.g., the parametric ReLU\cite{para_relu}. However, our adaptive matrix-valued activation functions are shown to be competitive and more robust in those experiments.

\subsection{Preliminaries}
For the simplicity of presentation, we consider a simple model data-fitting problem.  Given a training set containing 
$\{(x_n,f_n)\}_{n=1}^N$, where the coordinates $\{x_n\}_{n=1}^N\subset\mathbb{R}^d$ are the inputs, $\{f_n\}_{n=1}^N\subset\mathbb{R}^J$ are the output. They are implicitly related via an unknown target function $f:\mathbb{R}^d\to\mathbb{R}^J$ with the assumption that $f_n=f(x_n)$. 
The ReLU activation function is a piecewise linear function given by
\begin{equation*}\label{ReLU}
  \sigma(t)=\max\left\{t,0\right\},
  \quad \mbox{for} \quad t\in \mathbb{R}.
\end{equation*}
In the literature  $\sigma$ is acting component-wise on  an input vector. In a DNN, let $L$ be the
  number of layers and $n_\ell$ denote the number of neurons at the $\ell$-th layer for $0\leq\ell\leq L$ with $n_0=d$ and $n_L=J$. Let 
  $\mathcal{W}=(W_1,W_2,\ldots,W_L)\in\prod_{\ell=1}^L\mathbb{R}^{n_{\ell}\times n_{\ell-1}}$ denote the tuple of admissible weight matrices and  $\mathcal{B}=(b_1,b_2,\ldots,b_L)\in\prod_{\ell=1}^{L}\mathbb{R}^{n_\ell}$ the tuple of admissible bias vectors. The ReLU DNN approximation to $f$ at the $\ell$-th layer is recursively defined as
\begin{eqnarray}\label{DNN}
&&  \eta_\ell(x):=\sigma(W_{\ell} \eta_{\ell-1}(x) + b_{\ell})\in\mathbb{R}^{n_\ell},\quad \eta_0(x)=x\in
\mathbb{R}^d.
\end{eqnarray}
The traditional training process for such a DNN  is to find optimal
$\mathcal{W}_*\in\prod_{\ell=1}^L\mathbb{R}^{n_{\ell}\times n_{\ell-1}}$,  $\mathcal{B}_*\in\prod_{\ell=1}^{L}\mathbb{R}^{n_\ell}$, (and thus optimal $\eta_L=\eta_{L,\mathcal{W}_*,\mathcal{B}_*}$) such that
\begin{equation}\label{bestfit}
  (\mathcal{W}_*,\mathcal{B}_*)=\arg\min_{\mathcal{W}, \mathcal{B}}
  E(\mathcal{W},\mathcal{B}), \quad\mbox{where}\quad
E(\mathcal{W},\mathcal{B})=\sum_{n=1}^N \left|f_n-\eta_{L,\mathcal{W},\mathcal{B}}(x_n)\right|^2.
\end{equation}
In other words, $\eta_{L,\mathcal{W}_*,\mathcal{B}_*}$ best fits the data with respect to the discrete $\ell^2$-norm within the function class $\{\eta_{L,\mathcal{W},\mathcal{B}}\}$. In practice, the sum of squares norm in $E$ could be replaced with more convenient norms.

\section{Trainable matrix-valued activation function}

Having a closer look at ReLU $\sigma,$ we have a simple but quite useful observation that the activation $\sigma(\xi_\ell(x))$ with $\xi_\ell:=W_\ell\eta_{\ell-1} + b_\ell$ could be written as a matrix-vector multiplication $\sigma(\xi_\ell(x))=D_\ell(\xi_\ell(x))\xi_\ell(x)$,
where $D_\ell$ is a \emph{diagonal} matrix-valued function mapping from $\mathbb{R}^{n_\ell}$ to $\mathbb{R}^{n_\ell\times n_\ell}$ with diagonal being $\mathds{1}_{\{(0,\infty)\}}(s)$, thus taking values from the discrete set $\{0,1\}$. 
There is no reason to restrict on $\{0,1\}$ and we thus look for a larger set of values over which the diagonal entries of $D_\ell$ are running or sampled. With slight abuse of notation, our new DNN approximation to $f$ is calculated using the following recurrence relation
\begin{equation}\label{newDNN}
  \eta_0(x)=x\in\mathbb{R}^d,\quad\xi_\ell(x)=W_\ell\eta_{\ell-1}(x) + b_\ell,\quad \eta_\ell= (D_\ell\circ\xi_\ell)\xi_\ell,\quad
  \ell=1,\ldots,L.
\end{equation}
Here each $D_\ell$ is diagonal and is of the form \begin{equation}\label{Dell}
 D_\ell(y) = \operatorname{diag}(\alpha_{\ell,1}(y_1),\alpha_{\ell,2}(y_2),\ldots,\alpha_{\ell,n_\ell}(y_{n_\ell})),\quad y\in\mathbb{R}^{n_\ell},	
 \end{equation}
 where $\alpha_{\ell,i}(y_i)$ is a nonlinear function to be determined. Since piecewise constant functions can approximate a continuous function within arbitrarily high accuracy, we specify $\alpha_{\ell,i}$ with $1\leq i\leq n_\ell$ as
 \begin{equation}\label{alphali}
 	\alpha_{\ell,i}(s) = \left\{
\begin{aligned}
t_{\ell,i,0},     && s \in (-\infty, s_{\ell,i,1}], \\
t_{\ell,i,1},   && s \in (s_{\ell,i,1}, s_{\ell,i,2}],\\
\vdots \\
t_{\ell,i,m_{\ell,i}-1},  && s \in (s_{\ell,i,m_{\ell,i}-1}, s_{\ell,i,m_{\ell,i}}], \\
t_{\ell,i,m_{\ell,i}},  && s \in (s_{\ell,i,m_{\ell,i}}, \infty), 
\end{aligned}
\right.
\end{equation}
where $m_{\ell,i}$ is a positive integer and $\{t_{\ell,i,j}\}_{j=0}^{m_{\ell,i}}$ and $\{s_{\ell,i,j}\}_{j=1}^{m_{\ell,i}}$ are constants. We may suppress the indices $\ell,i$ in $\alpha_{\ell,i}$, $m_{\ell,i}$, $t_{\ell,i,j}$, $s_{\ell,i,j}$ and write them as $\alpha$, $m$, $t_j$, $s_j$ when those quantities are uniform across layers and neurons.
If $m=1$, $s_1=0$, $t_0=0$, $t_1=1,$ then the DNN \eqref{newDNN} is exactly the ReLU DNN.
If $m=1$, $s_1=0$, $t_1=1$ and $t_0$ is a fixed small negative number, \eqref{newDNN} reduces to the DNN based on Leaky ReLU. If $m=2$, $s_1=0$, $s_2=1$, $t_0=t_2=0$, $t_1=1$, then $\alpha=\alpha_{\ell,i}$ actually represents a discontinuous activation function. 

In our case, we shall fix some parameters from $\cup_{\ell=1}^L\cup_{i=1}^{n_\ell}\{t_{\ell,i,j}\}_{j=0}^{m_{\ell,i}}$ and $\cup_{\ell=1}^L\cup_{i=1}^{n_\ell}\{s_{\ell,i,j}\}_{j=1}^{m_{\ell,i}}$ and let the rest of them vary in the training process. When the diagonal cutoffs are fixed, while making the slopes learnable, this replicates the layer-wise adaptive rate scaling in \cite{you2017large}. 
Heuristically speaking, the activation functions among different layers in the resulting DNN may adapt to the target function $f$.
Since the nonparametric ReLU and a 1-parameter Leaky ReLU are special cases of the new activation functions, the proposed DNN with the new activations theoretically should bear at least the same approximation property. If the optimization problem is solved exactly, in practice the training error should be no worse than before. 
In the following, the activation function in \eqref{alphali} with trainable parameters in \eqref{newDNN} is named as ``trainable matrix-valued activation function (TMAF)".


Starting from the diagonal activation $D_\ell$, one can step further to construct more general activation matrices. First we note that $D_\ell$ could be viewed as a nonlinear operator $T_\ell: [C(\mathbb{R}^d)]^{n_\ell}\rightarrow [C(\mathbb{R}^d)]^{n_\ell}$, where
\begin{equation*}
    [T_\ell(g)](x)=D_\ell(g(x))g(x),\quad g\in [C(\mathbb{R}^d)]^{n_\ell},\quad x\in\mathbb{R}^d.
\end{equation*}
With this observation, one can parametrize a trainable nonlinear activation \emph{operator} determined by more general matrices, 
e.g., the following tri-diagonal operator \begin{equation}\label{Tell}
    [T_\ell(g)](x)=\begin{pmatrix}\alpha_{\ell,1}&\beta_{\ell,2}&0&\cdots&0\\
    \gamma_{\ell,1}&\alpha_{\ell,2}&\beta_{\ell,3}&\cdots&0\\
    \vdots&\ddots&\ddots&\ddots&\vdots\\
    0&0&\cdots&\alpha_{\ell,n_\ell-1}&\beta_{\ell,n_\ell}\\
    0&0&\cdots&\gamma_{\ell,n_\ell-1}&\alpha_{\ell,n_\ell}\end{pmatrix}g(x),\quad x\in\mathbb{R}^d.
\end{equation} 
The diagonal $\{\alpha_{\ell,i}\}$ is given in \eqref{alphali} while the off-diagonals $\beta_{\ell,i}$, $\gamma_{\ell,i}$ are piecewise constant functions in the $i$-th coordinate $y_i$ of $y\in\mathbb{R}^{n_\ell}$ defined in a fashion similar to $\alpha_{\ell,i}$. 
Theoretically speaking, even trainable full matrix activation is possible despite potentially increased training cost. In summary, 
the corresponding DNN based on trainable nonlinear activation operators $\{T_\ell\}_{\ell=1}^L$ reads
\begin{equation}\label{newDNN2}
  \eta_0(x)=x\in\mathbb{R}^d, \quad\xi_\ell(x)=W_\ell\eta_{\ell-1}(x) + b_\ell,\quad \eta_\ell:= T_\ell(\xi_\ell),\quad
  \ell=1,\ldots,L.
\end{equation}
 
The evaluation of $D_\ell$ and $T_\ell$ are cheap because they require only scalar multiplications and comparisons. When calling a general-purpose packages such as PyTorch in the training process, it is observed that the computational time of $D_\ell$ and $T_\ell$ is comparable to the classical ReLU. 
 
\begin{remark}
Our observation also applies to an activation function $\sigma$ other than ReLU. For example, we may rescale $\sigma(x)$ to obtain $\sigma(\omega_{i,\ell}x)$ for constants  $\{\omega_{i,\ell}\}$ varying layer by layer and neuron by neuron. Then $\sigma(\omega_{i,\ell}x)$ are used to form a matrix activation function and a TMAF DNN, where $\{\omega_{i,\ell}\}$ are trained according to given data and are adapted to the target function. This observation may be useful for specific applications. 
\end{remark}

\begin{remark}
We also note that the diagonal activation with a fixed bandwidth bear similarity with the piecewise-linear activation after the convolution layer in a so-called ADMM-net~\cite{yang2017admm}, yet is designed with fundamentally different philosophy. In \cite{yang2017admm}, an activation matrix with the size of the filter is composed the convolution operation. If one views the convolution operation as a diagonal matrix-valued operator, for a $p\times p$ convolution filter, the activation is a $n^2\times n^2$ diagonal matrix with the bandwidth being $p^2$. The nonlinearity in ADMM couples neighboring pixels in an image, while here TMAF does a nonlinear mixing in the channel dimension.
\end{remark}

\section{Numerical results}
In this section, we demonstrate the  feasibility  and efficiency of  TMAF by comparing it with the traditional ReLU-type activation functions. In principle, all parameters in \eqref{alphali} are allowed to be trained while we shall fix the intervals in \eqref{alphali} and only let function values $\{t_j\}$ vary for simplicity in the following. In each experiment, we use the same neural network structure, as well as the same learning rates, stochastic gradient descent (SGD)   optimization  and number NE of epochs (SGD iterations). In particular, the learning rate 1e-4 is used for epochs $1$ to $\frac{\rm NE}{2}$ and 1e-5 is used for epochs $\frac{\rm NE}{2}+1$ to ${\rm NE}$. 


\subsection{Function approximation (regression) problem}

For the first class of examples we use the $\ell^2$-loss function as defined in~\eqref{bestfit}.  For the classification problems we consider the \emph{cross-entropy} that is  widely used  as a loss function in classification models. The cross entropy is defined using a training set which consists of $p$ images, each with $N$ pixels. Thus, we have a matrix $Z\in \mathbb{R}^{N \times p}$ and each column corresponds to an image with $N$ pixels. Each image belongs to a fixed class $c_j$ from the set of image classes $\{c_k\}_{k=1}^p$, where $c_j\in \{1,\ldots,M\}$. The network structure maps $Z\in \mathbb{R}^{N\times p}$ to $X\in \mathbb{R}^{M\times p}$, and each column $x_j$ of $X$ is an output of the network evaluation at the corresponding column $z_j$ of $Z$. More precisely, 
\begin{equation*}\label{z-and-x}
  \begin{aligned}
& Z=(z_1,\ldots,z_p), \quad   X=(x_1,\ldots,x_p),\quad c_j=\operatorname{class}(z_j),\\
& x_j := \eta_{L,\mathcal{W},\mathcal{B}}(z_j), \quad  x_j\in \mathbb{R}^M, \quad z_j\in \mathbb{R}^N, \quad j=1,\ldots,p.
  \end{aligned}
\end{equation*}
The cross entropy loss function then is defined by
\begin{equation*}
  \begin{aligned}
 &\mathcal{C}(\mathcal{W},\mathcal{B})  = \sum_{k=1}^p -\log\left(\frac{\exp(x_{c_k,k})}{\sum_{j=1}^M\exp(x_{j,k})}\right),\\
  & (\mathcal{W}_*,\mathcal{B}_*) = \arg\max_{\mathcal{W},\mathcal{B}} \mathcal{C}(\mathcal{W},\mathcal{B}).
  \end{aligned}
\end{equation*}
To evaluate the loss function at a given image $z\in \mathbb{R}^N$, we first evaluate the network at $z$ with the given
$(\mathcal{W},\mathcal{B})=(\mathcal{W}_*,\mathcal{B}_*)$.
We then define the class $c(z)$ of $z$ and the loss $\operatorname{loss}(z)$ at $z$ as follows:
\begin{equation*}\label{loss-z}
  \begin{aligned}
  & \operatorname{c}(z)=\arg\max_{1\le j \le M}\left\{\exp(a_j)\right\}, \quad\mbox{where}\quad a=\eta_{L,\mathcal{W}_*,\mathcal{B}_*}(z), \\
   & \operatorname{loss}(z) = -\log\left(\frac{\exp(a_{c(z)})}{\sum_{j=1}^M\exp(a_{j})}\right).
  \end{aligned}
\end{equation*}
\subsubsection{Approximation of a smooth function}
\begin{figure}[!htbp]
  \centering
  \subfloat[$n=1$]{\includegraphics[scale=0.45]{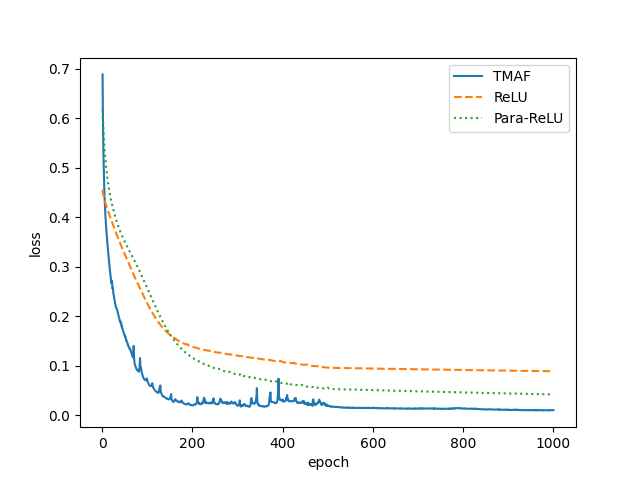}\label{fig:u1-1}}
  \hspace{0.1in}
  \subfloat[$n=2$]{\includegraphics[scale=0.45]{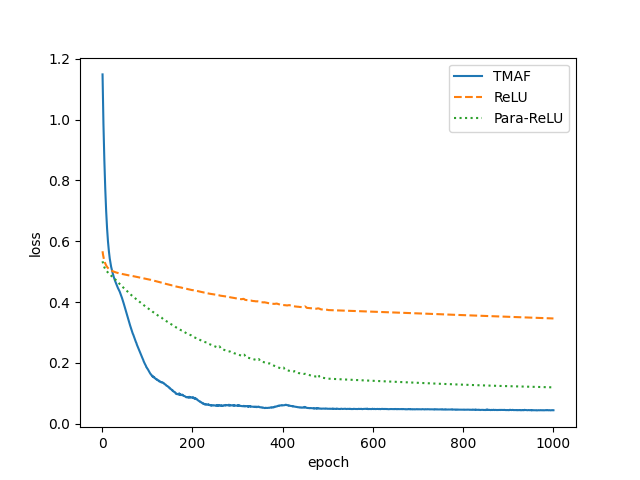}\label{fig:u1-2}}
  \caption{Training errors for $\sin(\pi x_{1}+ \cdots+\pi x_n)$, single hidden layer\label{fig:u1}}
\end{figure}

\begin{figure}[!htbp]
\centering
\subfloat[ReLU]{\includegraphics[scale=0.45]{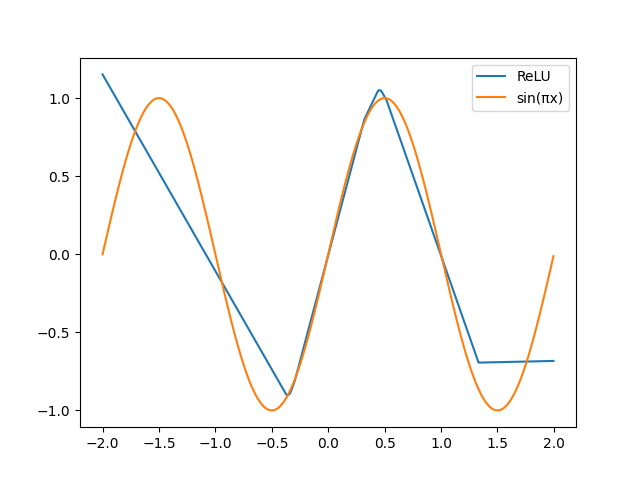}\label{fig:u4-1}}
\hspace{0.1in}
\subfloat[TMAF]{\includegraphics[scale=0.45]{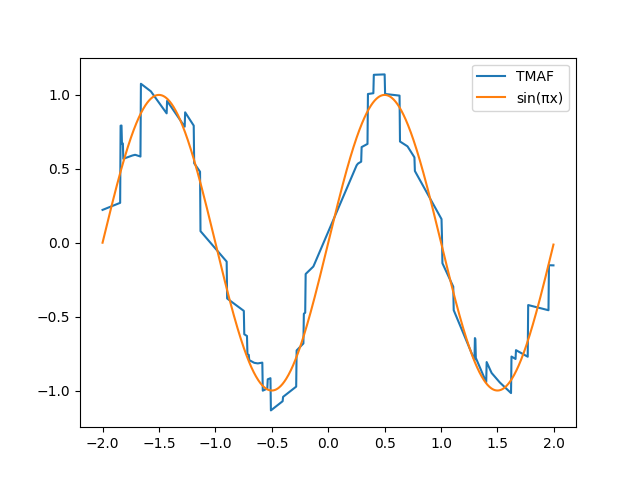}\label{fig:u4-2}}
\caption{Neural network approximations to $\sin(\pi x)$ , single hidden layer}
\label{fig:u4}
\end{figure}

\begin{figure}[!htbp]
\centering
\subfloat[$n=5$]{\includegraphics[scale=0.45]{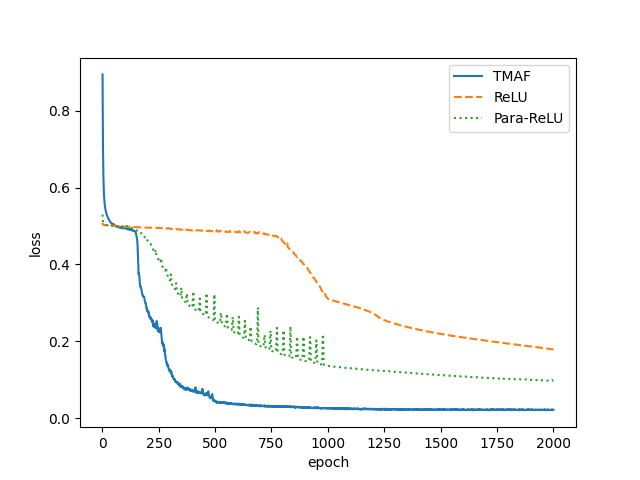}\label{fig:u1-5}}
\subfloat[$n=6$]{\includegraphics[scale=0.45]{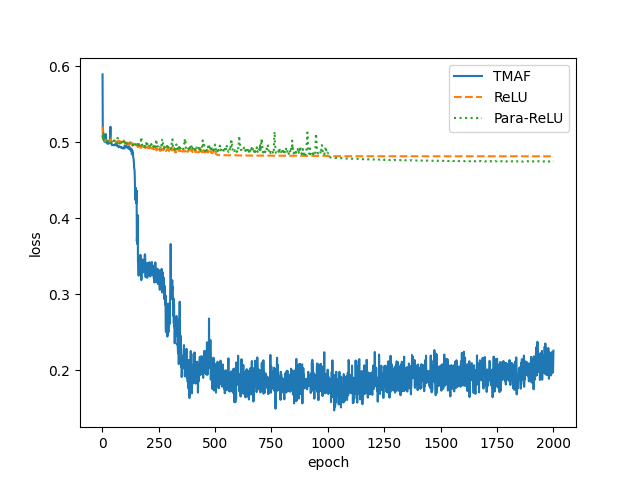}\label{fig:u1-6}}
\caption{Training errors for $\sin(\pi x_{1}+ \cdots+\pi x_n)$, two hidden layers.\label{fig:u56}}
\end{figure}

As our first example, we use neural networks to approximate
\begin{equation*}
  f(x_1,\cdots,x_n) = \sin(\pi x_1+ \cdots+\pi x_n),
  \quad x_k \in [-2,2], \quad k=1,\ldots,n.
\end{equation*}
The training datasets are 20000 input-output data pairs where the input data are randomly sampled from the hypercube $[-2,2]^n$. 
The networks \eqref{DNN} and \eqref{newDNN} have single or double hidden layers with $20$ neurons per layer. For TMAF $D_\ell$ in  \eqref{Dell}, the function $\alpha=\alpha_{\ell,i}$ in \eqref{alphali} uses intervals $(-\infty,-5)$, $(-5+k,-4+k]$, $(5,\infty)$, $0\leq k\leq9$. 
The approximation results are shown in
 Table~\ref{tab:2} and Figure~\ref{fig:u1}--\ref{fig:u56}. It is observed that TMAF is the most accurate activation approach. Moreover, the parametric ReLU does not approximate $\sin(\pi x_1+\ldots+\pi x_6)$ well,  see Figure \ref{fig:u1-6}. 
 \begin{table}
 \begin{tabular}{|l|l|l|l|l|l|l|l|l|}
 \hline
   & \multicolumn{8}{|c|}{Approximation error} \\ \cline{2-9}
   & \multicolumn{4}{|c|}{Single hiden layer} & \multicolumn{4}{|c|}{Two hiden layers}\\ \hline
   $n$   &      1 &     2 & 3    & 4    & 5 & 6 & 7 & 8 \\ \hline
   ReLU  & 0.089  & 0.34  & 0.39 & 0.41 
         & 0.14  & 0.21 & 0.25 & 0.31 \\ \hline
   TMAF  & 0.015 & 0.016 &  0.13 &  0.18
         & 0.07  & 0.105 &  0.153 &  0.17 \\ \hline\hline
 \end{tabular}
 \caption{Approximation errors for  $\sin(\pi x_1+\cdots+\pi x_n)$ by neural networks\label{tab:2}}
\end{table}

\subsubsection{Approximation of an oscillatory multi-frequency function}
\begin{figure}[!htbp]
\centering
\subfloat[Exact oscillating function]{\includegraphics[scale=0.45]{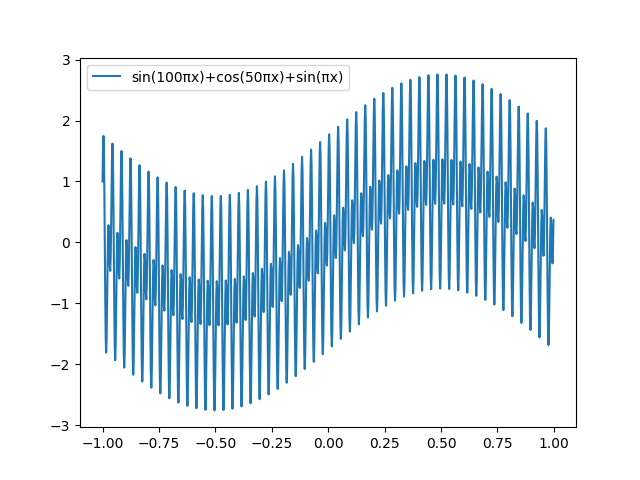}}
\hspace{0.1in}
\subfloat[Training loss comparison]{\includegraphics[scale=0.45]{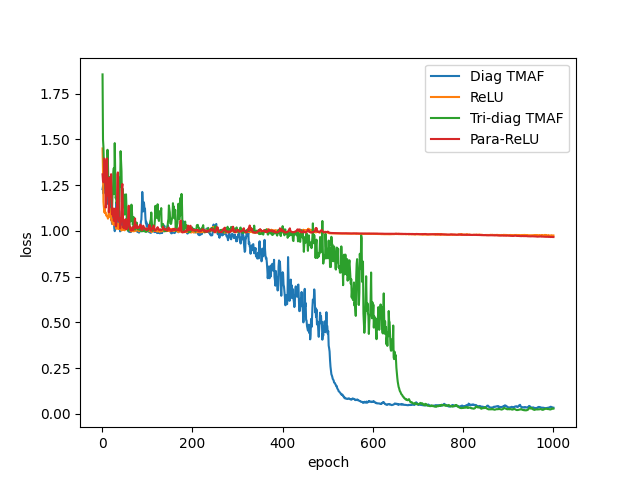}}
\caption{Plot of $f(x) = \sin(100 \pi x) + \cos(50 \pi x) + \sin(\pi x)$ and training loss comparison\label{complex_compare}}
\label{fig:u176}
\end{figure}

\begin{figure}[!htbp]
  \centering
\subfloat[ReLU approximation]{\includegraphics[scale=0.45]{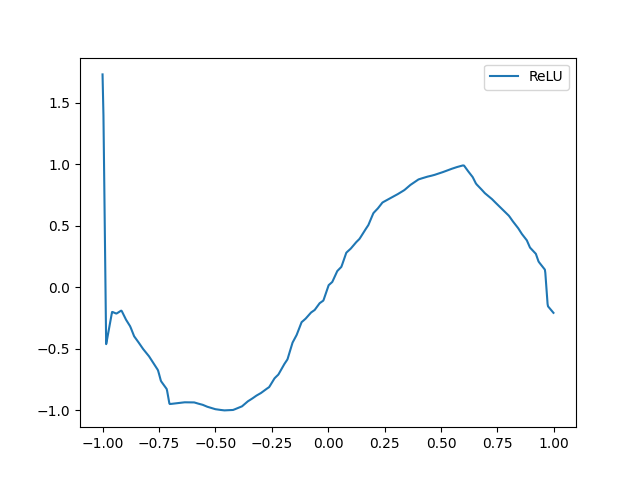}\label{fig:u9}}
\hspace{0.1in}
\subfloat[TMAF approximation]{\includegraphics[scale=0.45]{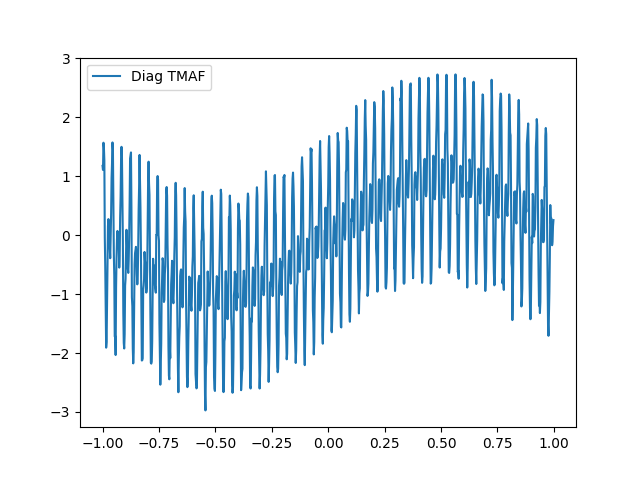}\label{fig:u10}}
\caption{Approximations to $f(x)=\sin(100 \pi x) + \cos (50 \pi x) + \sin(\pi x)$ by neural networks\label{fig:u11}}
\label{fig:u112}
\end{figure}

The next example is on approximating the following function having high and low frequency components
\begin{equation}\label{func}
  f(x) = \sin (100 \pi x) + \cos(50 \pi x) + \sin (\pi x),
\end{equation}
see Figure~\ref{complex_compare} for an illustration. The function in \eqref{func} is notoriously difficult to capture by numerical methods in scientific computing\SC{may need citation here}. In the context of approximation using NNs, it is observed in \cite{hong2022activation} that ReLU-based NN cannot resolve the high-frequency oscillatory feature of this function at all. The training datasets are 20000 input-output data pairs where the input data are randomly sampled from the interval $[-1,1]$. 
We test the diagonal TMAF \eqref{Dell} and the function $\alpha=\alpha_{\ell,i}$ \eqref{alphali} uses intervals $(-\infty,-5)$, $(-5+kh,-5+(k+1)h]$, $(5,\infty)$ with $h=0.1,$ $0\leq k\leq99$. We also consider the tri-diagonal TMAF \eqref{Tell}, where $\{\alpha_{\ell,i}\}$ is the same as the diagonal TMAF, $\{\beta_{\ell,i}\}$ and $\{\gamma_{\ell,i}\}$ are all piecewise constants based on intervals $(-\infty,-5+\underline{h})$, $(-5+kh+\underline{h},-5+(k+1)h+\underline{h}]$, $(5+\underline{h},\infty)$ and $(-\infty,-5+2\underline{h})$, $(-5+kh+2\underline{h},-5+(k+1)h+2\underline{h}]$, $(5+2\underline{h},\infty)$ with $\underline{h}=0.1/3$, $0\leq k\leq99$,  respectively. Numerical results could be found in Figures~\ref{complex_compare}, \ref{fig:u11} and Table \ref{table3}. 

For this challenging problem, we note that the diagonal TMAF and tri-diagonal TMAF produce high-quality approximations while ReLU and parametric ReLU are not able to approximate the highly oscillating function within reasonable accuracy. It is observed from Figure \ref{fig:u112} that ReLU actually approximates the low frequency part of \eqref{func}. To capture the high frequency, ReLU clearly has to use more neurons and thus much  more weight and bias parameters. On the other hand, increasing the number of intervals in TMAF only lead to a few more training parameters.

 \begin{table}
 \begin{tabular}{|l|l|l|}
 \hline
  & Error \\ \hline
 RELU & 0.97 \\ \hline
 Diag-TMAF   & 0.033 \\ \hline
 Tri-diag TMAF & 0.029\\ \hline
 \end{tabular}
 \caption{Error comparison for $f(x)=\sin(100 \pi x) + \cos (50 \pi x) + \sin(\pi x)$}
 \label{table3}
\end{table}

\subsection{Classification problem of \texttt{MNIST} and \texttt{CIFAR-10} data sets}
We now test TMAF by classifying images in the \texttt{MNIST} and \texttt{CIFAR-10} data sets. 
For TMAF $D_\ell$ in  \eqref{Dell}, the function $\alpha=\alpha_{\ell,i}$ \eqref{alphali} uses intervals $(-\infty,-5)$, $(-5+k,-4+k]$, $(5,\infty)$ with $0\leq k\leq9$. 

\begin{figure}[!htbp]
\centering
\subfloat[Training loss comparison]{\includegraphics[scale=0.45]{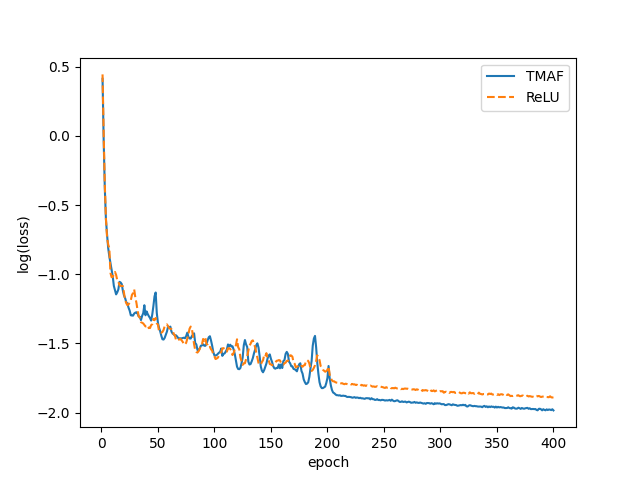}\label{u13}}
\hspace{0.1in}
\subfloat[Classification accuracy]
{\includegraphics[scale=0.45]{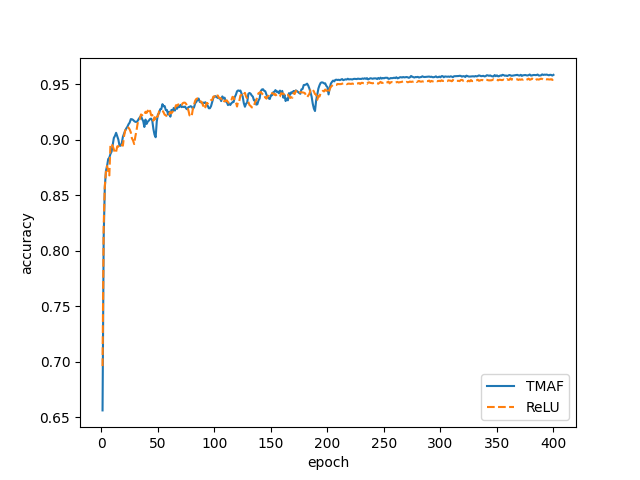}\label{u14}}
\caption{\texttt{MNIST}: Single hidden layer}
\label{u132}
\end{figure}

\begin{figure}[!htbp]
\centering
\subfloat[Training loss]{\includegraphics[scale=0.45]{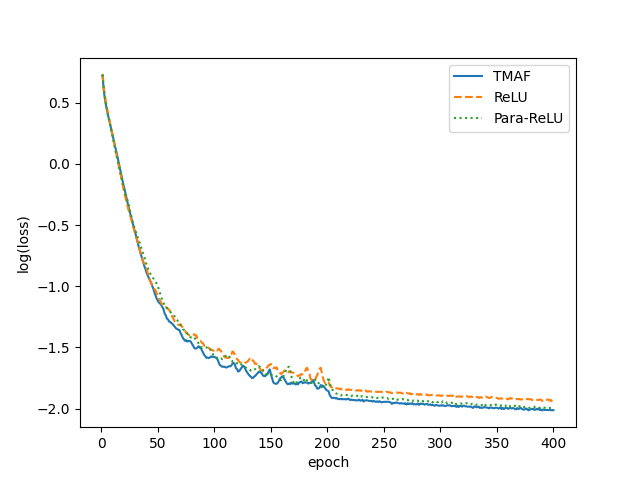}\label{u15}}
\hspace{0.1in}
\subfloat[Classification accuracy]
{\includegraphics[scale=0.45]{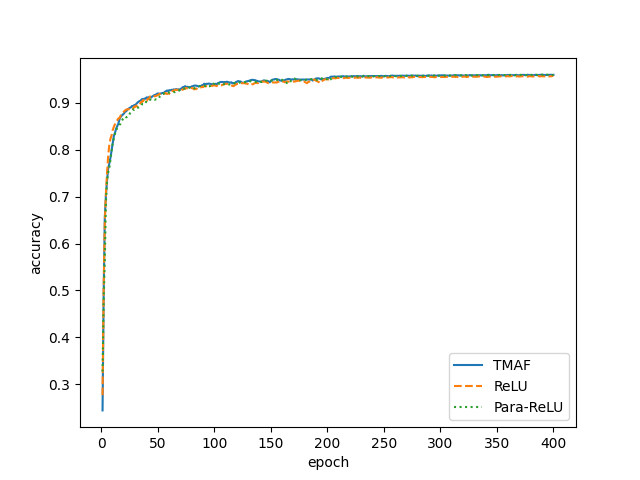}\label{u16}}
\caption{\texttt{MNIST}: Two hidden layers}
\end{figure}

For the \texttt{MNIST} set, we implement single and double layer fully connected networks \eqref{DNN} and \eqref{newDNN} with $10$ neurons per layer (except at the first layer $n_0=764$), and ReLU or diagonal TMAF \eqref{Dell} activation. 
Numerical results are shown in Figures~\ref{u13},~\ref{u14}, \ref{u15}, \ref{u16} and Table \ref{evaluation-mnist-cifar}. We note the the TMAF with single hidden layer ensures higher evaluation accuracy than ReLU, see Table  \ref{evaluation-mnist-cifar}. 


\begin{figure}[!htbp]
\centering
\subfloat[Training loss]{\includegraphics[scale=0.45]{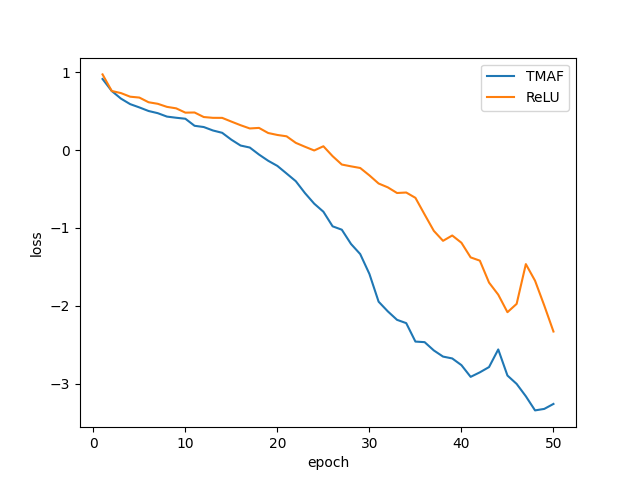}\label{u16-1}}
\hspace{0.1in}
\subfloat[Classification accuracy]{\includegraphics[scale=0.45]{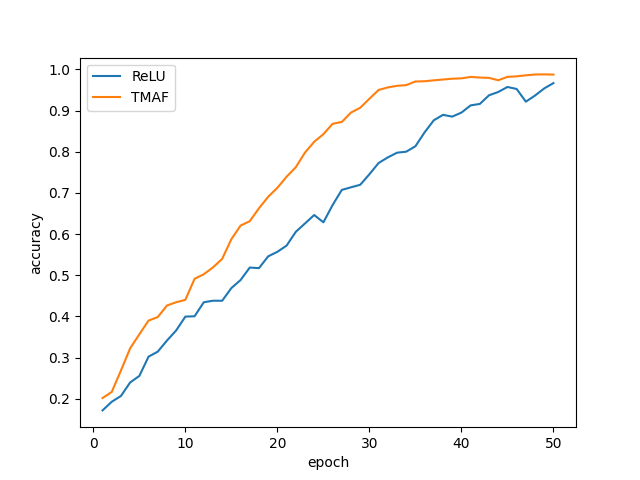}\label{u17}}
\caption{Comparison between ReLU and TMAF for \texttt{CIFAR-10}}
\end{figure}

For the \texttt{CIFAR-10} dataset, we use the \verb|ResNet18| network structure with $18$ layers and number of neurons provided by~\cite{HeZRS15}. The activation functions are still ReLU and the diagonal TMAF \eqref{Dell}. Numerical results are presented in
 Figures~\ref{u16-1}, ~\ref{u17} and  Table~\ref{evaluation-mnist-cifar}. Those parameters given in \cite{pytorch} are already tuned well with respect to ReLU. Nevertheless, TMAF still  produces  smaller errors in the training process and returns  better classification results in the evaluation stage. 

It is possible to improve the performance of TMAF applied to those benchmark datasets. The key point is to select suitable intervals in $\alpha_{\ell,i}$ to optimize the performance. A simple strategy is to let those intervals in \eqref{alphali} be varying and adjusted in the training process, which will be investigated in our future research.

\begin{table}
  \begin{tabular}{|l|l|l|}
 \hline\hline
  \multirow{2}{*}{Dataset} & \multicolumn{2}{|c|}{Evaluation Accuracy}     \\ \cline{2-3}
                                  & ReLU     & TMAF\\ \hline\hline
    \texttt{MNIST} (1 hidden layer) & 86.1\%  &  92.1\% \\ \hline
    \texttt{MNIST} (2 hidden layers) & 91.8\% & 92.2\%  \\ \hline
    \texttt{CIFAR-10} (Resnet18) & 92.8\%   & 93.2\%     \\ \hline\hline
 \end{tabular}
 \caption{Evaluation accuracy for the \texttt{MNIST} and \texttt{CIFAR-10}\label{evaluation-mnist-cifar}}

\end{table}

\bibliographystyle{alpha}
\bibliography{bib_tmaf/nn_b,bib_tmaf/pdo,bib_tmaf/transformers}

\end{document}